\documentclass[runningheads]{llncs}

\usepackage{eccv}

\usepackage{eccvabbrv}

\usepackage{graphicx}
\usepackage{booktabs}
\usepackage{multirow}

\usepackage{float}

\usepackage[accsupp]{axessibility}  

\usepackage[pagebackref,breaklinks,colorlinks,citecolor=eccvblue]{hyperref}
\usepackage{hyperref}

\usepackage{orcidlink}

\begin{document}

\title{Multimodal Taxonomic Conditioning for Generative Plankton Imagery} 

\titlerunning{Abbreviated paper title}

\author{Daniela Ivanova\inst{1}\orcidlink{0000-0002-3710-7413} \and
Özgü Göksu\inst{1,2}\orcidlink{0000-0002-5646-9946} \and
Nicolas Pugeault\inst{1}\orcidlink{0000-0002-3455-6280}}

\authorrunning{D.\ Ivanova et al.}
\institute{University of Glasgow, Glasgow, Scotland, UK \\
\email{Daniela.Ivanova@glasgow.ac.uk}, \email{Nicolas.Pugeault@glasgow.ac.uk} \and National Defence University, Istanbul, Türkiye \\
\email{ozgu.goksu@msu.edu.tr}
}

\maketitle

\begin{abstract}
  Automated plankton imaging produces severely long-tailed datasets, where the rare taxa of greatest ecological interest have too few images to train or evaluate classifiers reliably. 
  We generate synthetic plankton imagery conditioned on taxonomy: a CLIP encoder is adapted on a large plankton corpus with a ranked contrastive objective extended to deep, ragged taxonomies, then frozen to condition a parameter-efficient diffusion transformer. We evaluate synthetic sample quality on distributional fidelity and downstream classifier utility.
  \keywords{fine-grained recognition \and long-tailed recognition \and plankton}
\end{abstract}

\section{Introduction}
\label{sec:intro}

Automated instruments such as the Imaging FlowCytobot (IFCB)~\cite{olson2007submersible} generate plankton imagery far faster than experts can annotate it, and the resulting datasets are severely long-tailed: a handful of abundant taxa dominate, while ecologically important rare species may have fewer than ten images. Generative augmentation is an appealing remedy, but a demanding one for fine-grained morphology, where classes differ by details a generator must reproduce rather than approximate.

Taxonomy offers structure that ought to help. For instance, two \texttt{Chaetoceros} species share far more visual form than either shares with a ciliate; a generator that knows this can transfer from abundant relatives to rare ones. 
Recent work exploits this through conditional guidance~\cite{pan2025finediffusion} or through progressive rank-wise training~\cite{monsefi2025taxadiffusion} --- in both cases, the taxonomy enters through the training procedure rather than the class representation. We instead condition on embeddings whose geometry already encodes it, learned separately from generation: a CLIP encoder is adapted on a large plankton corpus with hierarchical rank annotations~\cite{montanares2026planktonzilla}, then frozen to condition a parameter-efficient diffusion transformer. Plankton annotation makes this harder than the fish and insect taxonomies such methods usually target: most specimens are identified only to genus or family, so the hierarchy is deep but unevenly populated, and ranked contrastive objectives do not apply directly.

Our contributions are: (i) an adaptation of ranked contrastive learning to deep, ragged biological taxonomies, with truncation-aware depth matching; (ii) a taxonomy-conditioned diffusion model for plankton that decouples hierarchical representation learning from generation; and (iii) an evaluation across distributional fidelity and downstream classifier utility.

\section{Related Work}
\label{sec:related}
\vspace{-0.8em}
\paragraph{Plankton recognition.} Kraft \etal~\cite{kraft2022towards} establish an operational pipeline for IFCB imagery classification, fine-tuning a ResNet-18 on 63K Baltic Sea images across 50 taxa and addressing class imbalance by random-oversampling rare classes to a minimum of 100 training images. Kareinen \etal~\cite{kareinen2025selfsupervised} exploit self-supervised pre-training on larger corpora of plankton images across imaging instruments and basins to further improve performance. Most recently, Planktonzilla-17M~\cite{montanares2026planktonzilla} consolidates 3.74M plankton images spanning 602 taxa and thirteen imaging systems, with  standardised taxonomic annotations across 7 taxonomic ranks. The authors show that CLIP~\cite{radford2021learning} trained with standard InfoNCE~\cite{oord2018representation} objective and taxonomic lineage as text is competitive with supervised classification, and outperforms even biological foundation models such as BioCLIP~\cite{stevens2024bioclip, gu2025bioclip}.

\paragraph{Fine-grained conditional generation.} Diffusion models produce high-fidelity images but struggle when classes are numerous and visually similar. FineDiffusion~\cite{pan2025finediffusion} scales class-conditional generation to 10{,}000 categories by fine-tuning only the class embedder, biases, and normalisation layers of a pretrained DiT~\cite{peebles2022scalable} --- $0.4\%$ of parameters --- and introduces hierarchical classifier-free guidance, replacing the unconditional branch with the sample's superclass so that guidance separates a class from its coarse neighbours rather than from noise. TaxaDiffusion~\cite{monsefi2025taxadiffusion} instead adapts Stable Diffusion~\cite{rombach2022highresolution} with LoRA~\cite{hu2021lora} modules trained progressively across taxonomic ranks, learning coarse levels before refining to species, and likewise substitutes a higher rank for the unconditional estimate at inference. Both acquire hierarchy through the training procedure or the guidance rule, while the class representation itself remains a learned embedding. 

\paragraph{Hierarchical contrastive learning.} Flat InfoNCE~\cite{oord2018representation}, as used by CLIP and Planktonzilla, treats every non-matching sample as equally negative, which discards the graded similarity a taxonomy provides. RINCE~\cite{hoffmann2022ranking} generalises InfoNCE to ranked positives, grading the denominator by rank so that more-similar classes are excluded from a given level's negatives. Its experiments use two ranks and assume every sample carries a label at every rank. Plankton taxonomies violate both assumptions: they are deep, and in practice \emph{ragged}, with many specimens identified only to genus or family. 

\vspace{-0.8em}
\section{Method}
\vspace{-1em}
\label{sec:method}
\paragraph{Our approach} differs from prior work in three ways. First, in how much of the hierarchy is used and how. FineDiffusion uses a single coarse level, and only in the guidance rule. TaxaDiffusion uses all ranks, but keeps CLIP frozen and learns per-rank refinement modules downstream of it, supervised only by the diffusion objective. Planktonzilla trains CLIP with the full lineage as a concatenated text string under a standard contrastive objective, so the taxonomy is present in the text but not in the objective. We instead form cumulative lineage strings at each rank and supervise their graded similarity directly, so the hierarchy is present in the embedding geometry before any generator sees it. Second, this adaptation uses a plankton corpus far larger than the generation target, letting the hierarchy be learned where the data supports it. Third, we adapt ranked contrastive learning to a deep, ragged taxonomy, with truncation-aware depth matching and per-rank weighting by batch coverage; to our knowledge the first such application.

We generate plankton images conditioned on taxonomy in two stages. First, a CLIP text encoder is adapted on a large plankton corpus with a rank-aware contrastive objective, producing per-class embeddings whose geometry reflects the taxonomic hierarchy. Downstream, these frozen multimodal embeddings replace the learned class-embedding table of a pretrained diffusion transformer.

\subsection{Taxonomy-aware text encoder}
\label{sec:encoder}
\paragraph{Setup.} We adapt OpenCLIP ViT-B/16 with LoRA~\cite{hu2021lora} ($r{=}\alpha{=}32$, applied to the last 12 $q, v, k$ blocks of both towers) on Planktonzilla-17M~\cite{montanares2026planktonzilla}, using the per-rank cumulative taxonomic lineage as text. Full CLIP training on this corpus takes 100 epochs on 64 $H100$ GPUs~\cite{montanares2026planktonzilla}. We use an effective batch size of $2048$ on two $A5000$ GPUs with a gradient accumulation step, learning rate of $1e-6$ with Adam, and OneCycle scheduler for 20 epochs.
\paragraph{Ranked contrastive learning.} Flat InfoNCE treats every non-matching class as equally negative, discarding the graded similarity a taxonomy provides. RINCE~\cite{hoffmann2022ranking} preserves this ordering by partitioning positives into ranks $P_1,\dots,P_R$ of decreasing similarity and applying InfoNCE recursively, treating coarser ranks as negatives at each level:
\begin{equation}
\ell_i = -\log \frac{\sum_{p \in P_i} \exp\!\big(h(q,p)/\tau_i\big)}{\sum_{p \in \bigcup_{j \geq i} P_j} \exp\!\big(h(q,p)/\tau_i\big) + \sum_{n \in N} \exp\!\big(h(q,n)/\tau_i\big)},
\label{eq:rince}
\end{equation}
with $\tau_i < \tau_{i+1}$, so that finer ranks are optimised at sharper temperatures. The total loss is $\sum_i \ell_i$.

\paragraph{Ragged taxonomies.} RINCE assumes every sample carries a label at every rank. Plankton annotations do not: often, specimens are identified only to genus or family. We define the rank of a pair by the shared depth of their lineages $\ell_i, \ell_j$ over $R$ taxonomic levels,
\begin{equation}
d(\ell_i, \ell_j) =
\begin{cases}
R, & \ell_i = \ell_j \\[2pt]
\max\{\, r : \ell_i^{1:r} = \ell_j^{1:r} \,\}, & \text{otherwise,}
\end{cases}
\label{eq:depth}
\end{equation}
where equality in the first case includes matching truncation -- shared depth otherwise counts only populated ranks.

\paragraph{Rank weighting.} A query has no rank-$d$ positive whenever no other batch member shares exactly that depth, and on ragged data the affected ranks differ sharply in how much of the batch they constrain. We normalise each rank by the batch size, so that a rank's influence is proportional to its coverage of the batch rather than uniform across ranks.

\subsection{Taxonomy-conditioned generation}
\label{sec:generation}
We follow FineDiffusion~\cite{pan2025finediffusion} in freezing a pretrained DiT-XL/2~\cite{peebles2022scalable} except for the conditioning embedder, biases, and normalisation layers ($2.5$M of $676$M parameters). The learned class-embeddings are replaced by the frozen CLIP encoder's embedding; a two-layer MLP projects it to the conditioning width and is the only new trainable component. The lineage text embedding is concatenated with the image embedding of the training image itself. We replace a fraction $p{=}0.5$ of training images' embeddings with their class mean. At inference, this enables conditioning on a class mean image embedding, as we do for our quantitative experiments, or on a specific image's embedding, as in Figure~\ref{fig:variations}. We retain hierarchical classifier-free guidance: during training a sample's species-level text embedding is replaced by the embedding of its \emph{phylum}-level lineage prefix with probability $0.1$. We use a batch size of $64$ across two $A5000$ GPUs throughout.


\section{Experiments}
\label{sec:experiments}
\paragraph{Dataset.} We use the WCO L4 Annotated IFCB Training Library~\cite{widdicombe2026}, collected at the Western Channel Observatory station L4 off Plymouth, UK: 74{,}181 IFCB images across 145 taxonomic classes, with a strogly long-tailed distribution. This collection is disjoint from Planktonzilla-17M, on which our text encoder is adapted, so the conditioning embeddings are transferred across collections rather than fitted to the generation target. We split 80/20 into train and test, then 90/10 again within train for validation, stratified by class following  Kraft \etal~\cite{kraft2022towards}, giving 52{,}749 / 6{,}595 / 14{,}837 images, with 58 training classes under 100 samples, and a single class left out of the test split due to stratification. The test split is real imagery throughout and identical across every experiment reported, whether the model was trained on real or generated data.

\paragraph{Generation.} Each model generates two sets. For the \emph{replacement} regime, one image per real training image, matching the real training and validation split class-for-class. For the \emph{augmentation} regime, top-ups for the 58 training classes below the 100-image threshold, $100 - n_c$ images per class $c$ (3{,}388 images in total). Sampling uses 250 DDPM steps at guidance scale $4.0$ for our models and the FineDiffusion baseline; TaxaDiffusion is sampled at the settings specified by its authors (250 steps, guidance scale $6$). All generated images are cropped to the detected organism with Grounding DINO~\cite{liu2024grounding} (prompt \texttt{"organism."}, box threshold $0.15$) before downstream classifier evaluation.

\subsection{Distributional fidelity}
\label{sec:fid}
We report FID~\cite{heusel2017gans} between each model's replacement set and the real training split, with all sets passed through an identical resize path. Generation at $256^2$ also compresses aspect ratios relative to real imagery (mean $1.19$ vs $1.40$), inflating all methods equally. As calibration, FID between two disjoint 3{,}000-image samples of real data is $10.98$ at that sample size.

\subsection{Downstream classifier utility}
\label{sec:classifier}
We test if generated images can substitute for or augment real training data. We adapt Kraft \etal~\cite{kraft2022towards}'s training and evaluation protocol: we train a frozen DINOv3~\cite{simeoni2025dinov} ViT-S/16 backbone with a three-layer linear head, 20 epochs, three seeds, evaluated by macro-averaged F1 score on the shared real test split. We evaluate training sets created both by the replacement (i.e. fully synthetic) and augmentation (rare classes topped up to 100 with generated images) regimes and we compare against naive oversampling by randomly duplicating real images~\cite{kraft2022towards}. 

\vspace{-1em}
\section{Results}
\vspace{-3em}
\begin{figure}[ht!]
    \centering
    \includegraphics[width=\textwidth]{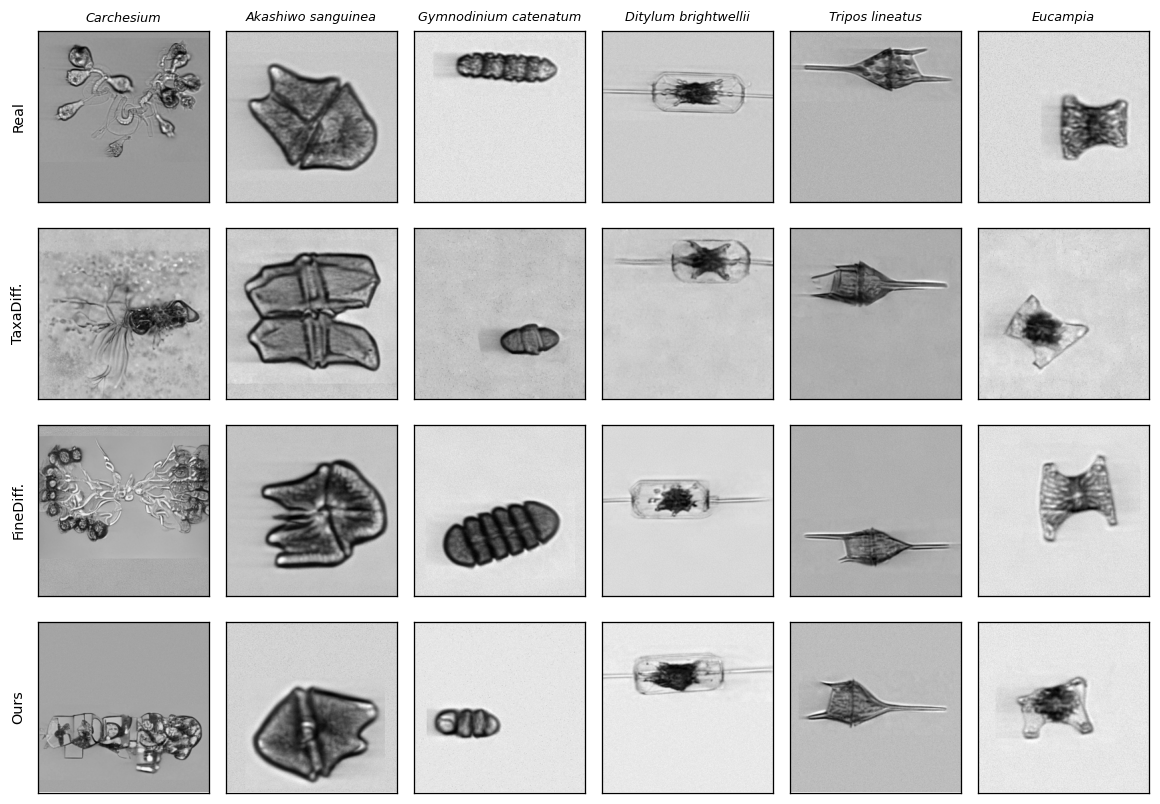}
    \caption{Qualitative comparison on three rare (left) and three common (right) classes, with training-set sizes of 2, 13, 7, 3591, 1492 and 3950 images respectively. Rows show a real specimen followed by one sample from each generator.}
    \label{fig:qualitative}
\end{figure}
\vspace{-1em}

Table~\ref{tab:classifier} summarises the results from our distributional fidelity and downstream classifier experiments. FineDiffusion is the better of the two baselines. We build upon it and replace its learned per-class embedding table with frozen CLIP text and image embeddings. Our taxonomic-informed conditioning improves fidelity (FID $19.17$ vs $22.43$) and the ability of synthetic data to substitute for real data, raising macro-$F1$ from $0.603$ to $0.664$ when the classifier is trained on generated images alone with the three generators ordered identically by FID and by replacement utility. When synthetic data instead only augments the rare classes of the real training set, the regime does not resolve differences between generators: ours, FineDiffusion and TaxaDiffusion are all statistically indistinguishable from duplicating real images, and duplication is itself indistinguishable from no oversampling at all ($0.875$ vs $0.869$). Our conditioning does underperform on the smallest classes, losing to duplication by $0.113$ macro-$F1$ over the four classes with $\leq 2$ training images. 

\vspace{-1em}
\begin{table}[]
\centering
\caption{Macro $F1$ over 144 classes on the shared real test split, mean $\pm$ standard deviation over three seeds, with the rare ($<100$ training images, $n{=}52$) and common ($n{=}92$) subsets shown separately. FID is measured between the same replacement set and the real training split. }
\label{tab:classifier}
\small
\begin{tabular}{llcccc}
\toprule
Regime & Training data & FID $\downarrow$ & F1 all $\uparrow$ & F1 rare $\uparrow$& F1 common $\uparrow$\\
\midrule
---        & real only              & ---   & $0.869 \pm 0.005$ & $0.786 \pm 0.014$ & $0.916 \pm 0.002$ \\
\midrule
\multirow{3}{*}{Replace}
           & TaxaDiffusion          & $43.62$          & $0.513 \pm 0.008$ & $0.305 \pm 0.015$ & $0.630 \pm 0.006$ \\
           & FineDiffusion          & $22.43$          & $0.603 \pm 0.003$ & $0.421 \pm 0.005$ & $0.706 \pm 0.001$ \\
           & Ours                   & $\mathbf{19.17}$ & $\mathbf{0.664 \pm 0.003}$ & $\mathbf{0.486 \pm 0.018}$ & $\mathbf{0.765 \pm 0.005}$ \\
\midrule
\multirow{4}{*}{Augment}
           & naive~\cite{kraft2022towards} & ---  & $0.875 \pm 0.003$ & $\mathbf{0.806 \pm 0.006}$ & $0.914 \pm 0.001$ \\
           & TaxaDiffusion          & ---              & $0.843 \pm 0.005$ & $0.718 \pm 0.014$ & $0.914 \pm 0.000$ \\
           & FineDiffusion          & ---              & $\mathbf{0.876 \pm 0.001}$ & $0.804 \pm 0.001$ & $\mathbf{0.917 \pm 0.001}$ \\
           & Ours                   & ---              & $0.867 \pm 0.003$ & $0.780 \pm 0.006$ & $0.915 \pm 0.002$ \\
\bottomrule
\end{tabular}
\vspace{-3em}
\end{table}
\begin{figure}[]
    \centering
    \includegraphics[width=\textwidth]{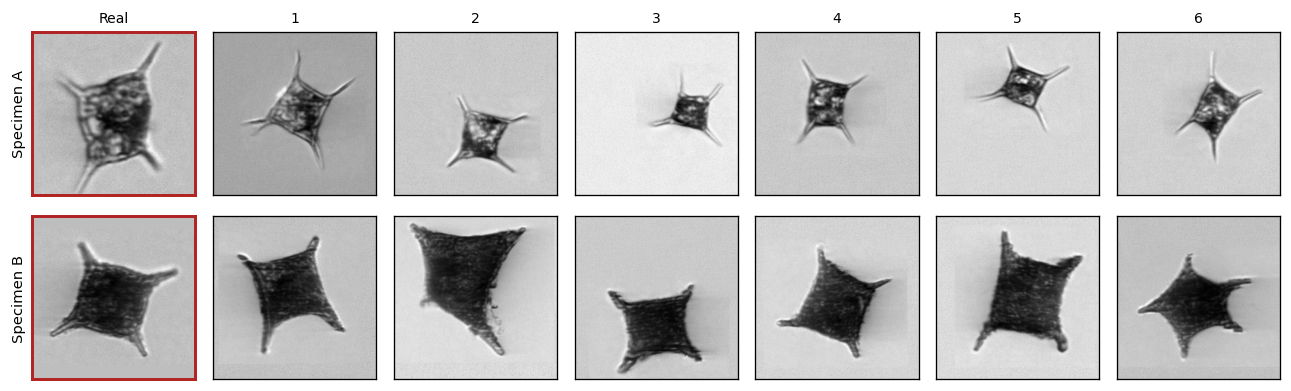}
    \caption{Specimen-conditioned generation. Each row shows a real training image (left) followed by four samples generated from that image's own CLIP image embedding rather than the class prototype.}
    \label{fig:variations}
\end{figure}
\vspace{-1em}

A possible explanation is that taxonomic proximity does not imply visual proximity, as with \textit{Carchesium} (Fig~\ref{fig:qualitative}, col. 1), a sessile colonial ciliate whose nearest taxonomic neighbours are free-swimming solitary ciliates and whose own images agree at $\cos=0.94$ in CLIP image space while its nearest neighbour by text embedding sits at $\cos=0.58$ in appearance. A frozen semantic embedding has nothing to transfer from in such cases, whereas a free per-class table can memorise the two available images.  Still, our approach presents a different and useful capability a learned per-class embedding table cannot express. Conditioning on a per-specimen CLIP image embedding rather than a class prototype generates variations of a particular individual, as in Fig.~\ref{fig:variations}. Our results motivate further exploration into modeling taxonomic relationships across modalities as conditioning signals for generative models.

\bibliographystyle{splncs04}
\bibliography{main}

@String(CVPR  = {IEEE Conf. Comput. Vis. Pattern Recog.})

@String(CVPRW = {IEEE Conf. Comput. Vis. Pattern Recog. Worksh.})

@String(AAAI  = {AAAI})

@String(CVPR  = {CVPR})

@String(CVPRW = {CVPRW})

@dataset{widdicombe2026,
  author    = {Widdicombe, Claire},
  title     = {{WCO L4 Annotated IFCB Training Library - Western English Channel, UK}},
  year      = {2026},
  version   = {1.2},
  publisher = {Zenodo},
  doi       = {10.5281/zenodo.20412178},
  url       = {https://doi.org/10.5281/zenodo.20412178}
}

@article{kraft2022towards,
  title={Towards operational phytoplankton recognition with automated high-throughput imaging, near-real-time data processing, and convolutional neural networks},
  author={Kraft, Kaisa and Velhonoja, Otso and Eerola, Tuomas and Suikkanen, Sanna and Tamminen, Timo and Haraguchi, Lumi and Yl{\"o}stalo, Pasi and Kielosto, Sami and Johansson, Milla and Lensu, Lasse and others},
  journal={Frontiers in Marine Science},
  volume={9},
  pages={867695},
  year={2022},
  publisher={Frontiers Media SA}
}

@article{montanares2026planktonzilla,
	author = {Montanares, Alan Gerson Contreras and Valenzuela, Luis and Mart{\' i}, Luis and S'anchez-Pi, N.},
	journal = {arXiv},
	year = {2026},
	pages = {},
	publisher = {},
	title = {Planktonzilla: Multimodal dataset and models for understanding plankton ecosystems},
	volume = {},
}

@inproceedings{kareinen2025selfsupervised,
	author = {Kareinen, Joona and Eerola, Tuomas and Kraft, Kaisa and Lensu, Lasse and Suikkanen, Sanna and K{\" a}lvi{\" a}inen, Heikki},
	booktitle = {2025 {IEEE}/{CVF} {Conference} on {Computer} {Vision} and {Pattern} {Recognition} {Workshops} ({CVPRW})},
	year = {2025},
	pages = {2122--2132},
	organization = {IEEE},
	title = {Self-{Supervised} {Pretraining} for {Fine}-{Grained} {Plankton} {Recognition}},
	volume = {},
}

@inproceedings{gu2025bioclip,
	author = {Gu, Jianyang and Stevens, Samuel and Campolongo, Elizabeth G and Thompson, Matthew J and Zhang, Net and Wu, Jiaman and Kopanev, Andrei and Mai, Zheda and White, Alexander E. and Balhoff, James and Dahdul, Wasla and Rubenstein, Daniel and Lapp, Hilmar and Berger-Wolf, Tanya and Chao, Wei-Lun and Su, Yu},
	booktitle = {The {Thirty}-ninth {Annual} {Conference} on {Neural} {Information} {Processing} {Systems}},
	year = {2025},
	pages = {},
	organization = {},
	title = {BioCLIP 2: Emergent {Properties} from {Scaling} {Hierarchical} {Contrastive} {Learning}},
	volume = {},
}

@inproceedings{stevens2024bioclip,
	author = {Stevens, Samuel and Wu, Jiaman and Thompson, Matthew J and Campolongo, Elizabeth G and Song, Chan Hee and Carlyn, David Edward and Dong, Li and Dahdul, Wasila M and Stewart, Charles and Berger-Wolf, Tanya and Chao, Wei-Lun and Su, Yu},
	booktitle = {IEEE/{CVF} {Conference} on {Computer} {Vision} and {Pattern} {Recognition} ({CVPR})},
	year = {2024},
	pages = {},
	organization = {},
	title = {BioCLIP: A {Vision} {Foundation} {Model} for the {Tree} of {Life}},
	volume = {},
}

@article{pan2025finediffusion,
	author = {Pan, Ziying and Wang, Kun and Li, Gang and He, Feihong and Lai, Yongxuan},
	journal = {Applied Intelligence},
	number = {4},
	year = {2025},
	pages = {309},
	publisher = {},
	title = {FineDiffusion: scaling up diffusion models for fine-grained image generation with 10,000 classes.},
	volume = {55},
}

@inproceedings{peebles2022scalable,
	author = {Peebles, William S. and Xie, Saining},
	booktitle = {IEEE {International} {Conference} on {Computer} {Vision}},
	year = {2022},
	pages = {4172--4182},
	organization = {},
	title = {Scalable {Diffusion} {Models} with {Transformers}},
	volume = {},
}

@inproceedings{monsefi2025taxadiffusion,
	author = {Monsefi, Amin Karimi and Khurana, Mridul and Ramnath, Rajiv and Karpatne, A. and Chao, Wei-Lun and Zhang, Cheng},
	booktitle = {IEEE {International} {Conference} on {Computer} {Vision}},
	year = {2025},
	pages = {8579--8589},
	organization = {},
	title = {Taxadiffusion: Progressively {Trained} {Diffusion} {Model} for {Fine}-{Grained} {Species} {Generation}},
	volume = {abs/2506.01923},
}

@inproceedings{rombach2022highresolution,
	author = {Rombach, Robin and Blattmann, Andreas and Lorenz, Dominik and Esser, Patrick and Ommer, Bjorn},
	booktitle = {2022 {IEEE}/{CVF} {Conference} on {Computer} {Vision} and {Pattern} {Recognition} ({CVPR})},
	year = {2022},
	pages = {10674--10685},
	organization = {IEEE},
	title = {High-{Resolution} {Image} {Synthesis} with {Latent} {Diffusion} {Models}},
	volume = {},
}

@inproceedings{hu2021lora,
	author = {Hu, J. and Shen, Yelong and Wallis, Phillip and Allen-Zhu, Zeyuan and Li, Yuanzhi and Wang, Shean and Chen, Weizhu},
	booktitle = {International {Conference} on {Learning} {Representations}},
	year = {2021},
	pages = {},
	organization = {},
	title = {LoRA: Low-{Rank} {Adaptation} of {Large} {Language} {Models}},
	volume = {abs/2106.09685},
}

@article{oord2018representation,
	author = {Oord, A{\" a}ron van den and Li, Yazhe and Vinyals, O.},
	journal = {arXiv.org},
	year = {2018},
	pages = {},
	publisher = {},
	title = {Representation {Learning} with {Contrastive} {Predictive} {Coding}},
	volume = {abs/1807.03748},
}

@inproceedings{hoffmann2022ranking,
	author = {Hoffmann, David T. and Behrmann, Nadine and Gall, Juergen and Brox, T. and Noroozi, M.},
	booktitle = {AAAI {Conference} on {Artificial} {Intelligence}},
	year = {2022},
	pages = {},
	organization = {},
	title = {Ranking {Info} {Noise} {Contrastive} {Estimation}: Boosting {Contrastive} {Learning} via {Ranked} {Positives}},
	volume = {abs/2201.11736},
}

@inproceedings{radford2021learning,
	author = {Radford, Alec and Kim, Jong Wook and Hallacy, Chris and Ramesh, A. and Goh, Gabriel and Agarwal, S. and Sastry, G. and Askell, Amanda and Mishkin, Pamela and Clark, Jack and Krueger, Gretchen and Sutskever, I.},
	booktitle = {International {Conference} on {Machine} {Learning}},
	year = {2021},
	pages = {},
	organization = {},
	title = {Learning {Transferable} {Visual} {Models} {From} {Natural} {Language} {Supervision}},
	volume = {abs/2103.00020},
}

@inproceedings{liu2024grounding,
  title={Grounding dino: Marrying dino with grounded pre-training for open-set object detection},
  author={Liu, Shilong and Zeng, Zhaoyang and Ren, Tianhe and Li, Feng and Zhang, Hao and Yang, Jie and Jiang, Qing and Li, Chunyuan and Yang, Jianwei and Su, Hang and others},
  booktitle={European conference on computer vision},
  pages={38--55},
  year={2024},
  organization={Springer}
}

@article{olson2007submersible,
  title={A submersible imaging-in-flow instrument to analyze nano-and microplankton: Imaging FlowCytobot},
  author={Olson, Robert J and Sosik, Heidi M},
  journal={Limnology and Oceanography: Methods},
  volume={5},
  number={6},
  pages={195--203},
  year={2007},
  publisher={Wiley Online Library}
}

@article{simeoni2025dinov,
	author = {Sim{\' e}oni, Oriane and Vo, Huy V. and Seitzer, Maximilian and Baldassarre, Federico and Oquab, Maxime and Jose, Cijo and Khalidov, Vasil and Szafraniec, Marc and Yi, Seungeun and Ramamonjisoa, Michael and Massa, Francisco and Haziza, Daniel and Wehrstedt, Luca and Wang, Jianyuan and Darcet, Timoth{\' e}e and Moutakanni, Th{\' e}o and Sentana, Leonel and Roberts, Claire and Vedaldi, Andrea and Tolan, Jamie and Brandt, John and Couprie, Camille and Mairal, J. and J{\' e}gou, Herv{\' e} and Labatut, Patrick and Bojanowski, Piotr},
	journal = {arXiv},
	year = {2025},
	pages = {},
	publisher = {},
	title = {DINOv3},
	volume = {},
}

@article{heusel2017gans,
  title={Gans trained by a two time-scale update rule converge to a local nash equilibrium},
  author={Heusel, Martin and Ramsauer, Hubert and Unterthiner, Thomas and Nessler, Bernhard and Hochreiter, Sepp},
  journal={Advances in neural information processing systems},
  volume={30},
  year={2017}
}
\end{document}